\documentclass[submission,copyright,creativecommons]{eptcs}

\newcommand{\n}{\neg}
\newcommand{\et}{\wedge}

\newcommand{\la}{\langle}
\newcommand{\ra}{\rangle}

\newcommand{\prcon}[3]{\mbox{$\mathtt{P}^{#3}(\mathtt{#1}\vert \mathtt{#2})$}}
\newcommand{\prtext}[2]{\mbox{$\mathtt{P}^{#2}(\mathtt{#1})$}}

\newcommand{\pr}[1]{\mbox{$\mathtt{P}(#1)$}}

 \newcommand{\df}[1]{\mbox{$\pr{#1}\downarrow$}}
 \newcommand{\ndf}[1]{\mbox{$\pr{#1}\uparrow$}}

\providecommand{\event}{TARK 2017} 
\usepackage{underscore}           

\usepackage{amsmath}
\usepackage{amssymb}
\usepackage{amsthm}

\title{Reconciling  Bayesian Epistemology and Narration-based Approaches to Judiciary Fact-finding}
\author{Rafal Urbaniak
\institute{Centre for Logic and Philosophy of Science, Ghent University, Belgium\thanks{The research has been funded by Research Foundation Flanders.}}
\institute{Institute of Philosophy, Sociology and Journalism, University of Gdansk, Poland\thanks{The research was supported by National Centre for Science grant number 2016/22/E/HS1/00304.}}
\email{rfl.urbaniak@gmail.com}}
\def\titlerunning{Bayesian Epistemology and Narratives}
\def\authorrunning{R. Urbaniak}
\begin{document}
\maketitle

\begin{abstract}
Legal probabilism (LP) claims the  degrees of conviction in juridical fact-finding  are to be modeled exactly the way degrees of beliefs are modeled in standard bayesian epistemology. Classical legal probabilism (CLP) adds that the conviction is justified if the credence in guilt given the evidence is above an appropriate guilt probability threshold.  The views are challenged on various counts, especially by the proponents of the so-called narrative approach, on which the fact-finders' decision is the result of a dynamic interplay between competing narratives of what happened.  I develop a way a bayesian epistemologist can make sense of the narrative approach. I do so by formulating a probabilistic framework for evaluating  competing narrations in terms of formal  explications of the informal evaluation criteria used in the narrative approach.\footnote{Due to the nature of this volume and space restrictions, this paper focuses mostly on presenting the technical apparatus. Deeper philosophical discussion, application to examples, and further informal clarifications are postponed to a different, lengthy  paper.}
\end{abstract}

\section{Legal probabilism and its troubles}

According to \emph{legal probabilism} (LP), degrees of conviction in juridical fact-finding are to be modeled exactly the way degrees of beliefs are modeled in standard bayesian epistemology: by means of probabilistic distributions satisfying the standard axioms of probability theory.
 \emph{Classical legal probabilism} (CLP), which   originated with Bernoulli \cite{Bernoulli1713Ars-conjectandi} adds on top of that the view according to which the criminal standard of proof beyond reasonable doubt should be equated with a certain high threshold probability  of guilt (albeit, some variants of the view admit that thresholds for different cases might be different).\footnote{Nowadays,  legal scholars fond of probabilism usually inscribe to LP but not to CLP.}

 LP (and generally, the use of probabilistic methods in judiciary contexts) is criticized  from various angles
 (see for example \cite{tribe1970further}, \cite{tribe1971trial},  \cite{Cohen1977The-probable-an}, \cite{Underwood1977The-thumb-on-th}, 
 \cite{Nesson1979Reasonable-doub}, 
 \cite{cohen1981subjective}, 
 \cite{dant1988gambling}, 
 \cite{wells1992naked},
 \cite{Stein2005Foundations-of-},
 \cite{allen2007problematic},
 \cite{ho2008philosophy},
 \cite{haack2011legal}).   The critics of LP  argue that the view is blind to various phenomena that an adequate philosophical  account of legal fact-finding should explain. Some of them pertain to   procedural issues \cite{Stein2005Foundations-of-}: proceedings are back-and-forth between opposing parties, cross-examination is crucial, and yet CLP seems to take no notice of this dynamics. Some have to do with reasoning methods which are not only evidence-to-hypothesis, but also hypotheses-to-evidence \cite{wells1992naked,allen2007problematic}  and involve  inference to the best explanation \cite{dant1988gambling}. A better account, arguably, is one in which the proceedings are seen as  an interplay of evidence and various explanations (often called \emph{narratives})  presented by opposing parties   \cite{ho2008philosophy}. 
 
 Accordingly, the no plausible alternative story (NPAS) theory \cite{allen2010no} is that the courtroom is a confrontation of competing narrations offered by the defendant and by the prosecutor and  the narrative to be selected should be the most plausible one. The view is conceptually plausible \cite{di2013statistics} and finds  support  in psychological evidence \cite{pennington1991cognitive,pennington1992explaining}.  The approach is also better at capturing the already listed  phenomena that CLP is claimed to be blind to.

From the perspective of a formal epistemologist,  the key disadvantage of NPAS is that it abandons the rich toolbox of probabilistic methods and takes the key notion of plausibility to be a primitive notion which should be intuitively  understood.\footnote{``I have been asked to elaborate on the meaning of “plausibility” in this theoretical structure. The difficulty in doing so is that the relative plausibility theory is a positive rather than a normative theory. For reasons I will elaborate below, “plausibility” can serve as a primitive theoretical term the meaning of which is determined by its context in the explanation of trials.'' \cite[p. 10]{allen2010no}} The goal of this paper is to develop a bayesian approach to NPAS, showing that one can embrace NPAS without giving up  on probabilistic methods.

 \section{Classical legal probabilism and the gatecrasher paradox}

CLP is also susceptible to criticism directed at its specific  claim: that there is a guilt probability threshold. One of the most well-known conceptual arguments against the idea is the following.

The \emph{paradox of the gatecrasher} \cite{Cohen1977The-probable-an}\footnote{It is analogous the \emph{prisoners in a yard scenario} \cite{Nesson1979Reasonable-doub}, where a group of prisoners commits a group killing, and it's impossible to identify the single innocent prisoner. Mathematically, the examples are pretty much the same.} has been developed  to indicate that   mere high probability of guilt is not enough for a conviction. A variant of the paradox goes as follows: 
\begin{quote}
Suppose our guilt threshold is high, say at 0.99. Consider the situation in which 1000 fans enter a football stadium, and 999 of them avoid paying for their tickets. A random spectator is tried for not paying. The  probability that the spectator under trial did not pay exceeds 0.99.   Yet, intuitively, a spectator cannot be considered guilty on the sole basis of the number of people who did and did not pay.
\end{quote}

 CLP fails to handle the gatecrasher paradox: whatever $\neq 1$ threshold you propose, I can give you a variant of the gatecrasher with sufficiently many people for the guilt statement of an arbitrary spectator to be above the guilt threshold. 

 NPAS, on the other hand, seems to avoid the paradox: conviction of a random attendee is unjustified, because there is no real plausible  story of guilt of that particular person. After all, merely quoting the statistics, on this approach, hardly counts as giving a narrative. The problem is, such considerations are left at an  informal level, the requirements on what should count as a narration aren't too clear, and so the considerations remain somewhat indecisive. After all, a stubborn philosopher might still insist: \emph{look, all that is needed is a theory of what happened that is more plausible than the alternative --- my theory is, the suspect is guilty, and, given the evidence, it is more plausible than the alternative, which says that he isn't.}

 \section{New legal probabilism}

\emph{New legal probabilism} (NLP)  is an attempt to improve on the underspecificity of NPAS \cite{di2013statistics}. While still being informal, the approach is more specific about the conditions that a successful accusing narration is to satisfy for the conviction beyond reasonable doubt to be justified. Di Bello identifies four key requirements.

 \begin{center}
\begin{tabular}{lp{8.5cm}}
 (Evidential support) &The defendant's guilt probability on the evidence should be sufficiently supported by the evidence, and a successful accusing narration should explain the relevant evidence. \\
(Evidential completeness) &  The evidence available at trial should be complete as far as a reasonable fact-finders' expectations are concerned. \\
(Resiliency)&  The prosecutor's narrative, based on the available evidence, should not be susceptible to revision given reasonably possible future arguments and evidence. \\
(Narrativity) & The narrative offered by the prosecutor should answer all 
the natural or reasonable questions one may have about what happened, given the content of the prosecutor's narration and the available evidence.
\end{tabular}\end{center}

	How would NLP handle the gatecrasher? Di Bello's discusses an analogous \emph{prisoners} scenario, in which a footage indicates that all but one prisoners participated in a group killing, but the footage doesn't allow for identification of the innocent prisoner \cite[pp. 219-222]{di2013statistics}. He argues that the conviction is unjustified on two counts. First, the narrative is grossly incomplete: ``What is the initiating event? What psychological responses did it trigger? Who participated in the killing? What were the different participants doing?\dots''   Second, the convicting scenario, Di Bello argues, isn't resilient, for it is quite plausible that more evidence might become available.

Unfortunately, it is not quite clear whether the resources that NLP helps herself to when formulating the requirements,  indeed fall into the realm of bayesian methodology:

\begin{quote}  
The probabilists can enrich their framework by adding probability-based accounts of evidential completeness, resiliency, and narrativity. To my knowledge, no legal probabilist has undertaken the task in any systematic way. \cite[p. 75]{di2013statistics}
\end{quote}

In what follows we undertake this task.

\section{Preliminary technicalities}

\subsection{The language and its interpretation}

The \emph{object language} is a standard propositional language 
$\mathcal{L}$ (I assume $\et$ and $\neg$ are the primitive connectives, nothing serious hinges on the choice) extended to a language $\mathcal{L}^{+}$ with primitive unary operators $E, N^A_1, \dots, N^A_k$, $N^D_1, \dots, N^D_k$, and the guilt statement constant $G$. 

The content of the guilt statement is given in terms of a list of conditions in the background language that need to be established for a conviction to be justified. This is modeled by conditioning on the definition of guilt  $\mathtt{G}$ which has the form of 
$G\equiv g_1\et \cdots \et g_l$
  for appropriate $g_1, \dots, g _l\in \mathcal{L}$.

The intended interpretation of $Ep$  is \emph{$p$ is part of evidence}. The idea is that 
after all the evidence and all the arguments have been presented in court, the background knowledge is to be  enriched by the  pieces of evidence presented, thought of as sentences of $\mathcal{L}$:   $\mathtt{E} = \{e_1,\dots, e_j\}\subset \mathcal{L}$. However, we are not only to extend our beliefs by $e_1,\dots, e_j$, but also by the corresponding claims about these sentences being part of evidence: $Ee_1, \dots, E e_j$.

 $N^A_ip$ means \emph{$p$ is part of an accusing narration $\mathtt{N}^A_i$} and
 $N^D_ip$ means \emph{$p$ is part of a defending narration $\mathtt{N}^D_i$}. 
 Each narration $\mathtt{N}_i$ (in contexts in which it is irrelevant whether a narration is  an accusing one or not, I will suppress the superscripts) is taken to be  a finite set of sentences $n_{i1}, n_{i2}, \dots, n_{ik_{i}}$ of $\mathcal{L}^+$. 

$\mathtt{E}$ stands ambiguously for the set of all sentences constituting evidence and for the conjunction thereof. Which reading is meant will always be clear from the context (this convention applies to all finite sets of sentences considered in this paper). $\mathtt{E}^d$ stands for  $\mathtt{E}^d=\{E\varphi \vert \varphi \in \mathtt{E}\}$ and 
 $\mathtt{E}^-$ is $\{\n E\varphi \vert \varphi \not \in \mathtt{E}\}$. The distinction is needed, because there is a difference between knowing that certain  sentences are pieces of evidence, and knowing that no  other sentence is.
 For any narration $\mathtt{N}_i$,  symbols $\mathtt{N}_i$, $\mathtt{N}_i^d$, and $\mathtt{N}_i^-$  are to be understood analogously to $\mathtt{E}$, $\mathtt{E}^d$ and $\mathtt{E}^-$. 
  $\mathtt{N}^d$ is the (positive) \emph{description} of all the narrations, $\bigcup_i\mathtt{N}_i^d$,  and $\mathtt{N}^-$= $\bigcup_i  \mathtt{N}^-_i$ adds that this description is complete.

\subsection{Partial probability and four  thresholds}

The current framework diverges from the standard bayesianism in using partial probability functions rather than full probabilistic discributions to model credences. This is motivated by noticing that   the fact-finders, on one hand, are supposed to rely on their background knowledge when assessing the plausibility of a given narration, but on the other hand, they  clearly cannot rely on all biases and assumptions that they have.\footnote{Suspending our conviction about $p$ cannot be easily modeled in the standard bayesian framework, since even the most sensible candidate, $1/2$, doesn't do the job. Just to give a simple example, there is a difference between knowing that a given coin is fair and assigning probability of $1/2$ to heads, and not knowing how fair a coin is at all and assigning probability of $1/2$ to heads for this reason.} 
 
A  \emph{ partial conditional credence function} $\mathtt{P}$ (partially) maps  $\mathcal{L^+}\times \mathcal{P}(\mathcal{L^+})$ to  $[0,1]$.\footnote{Partial credence functions for conditional probabilities have been introduced in  \cite{lepage2003probabilistic} and \cite{lepage2012partial}; my definition differs from that account in a few inessential aspects.}  (I often write $\pr{h\vert \mathtt{E}}$ instead of $\pr{\la h, \mathtt{E}\ra}$).     Let  $\downarrow$ and $\uparrow$ stand for \emph{being defined} and \emph{being undefined} respectively.   A partial probability distribution has to have  an extension to a total conditional probability distribution over $\mathcal{L^+}$  satisfying the standard axioms of conditional probability,  Moreover, it  additionally has to satisfy  the following conditions  for any $\Gamma \subseteq \mathcal{L}^+$, and any $\varphi, \psi\in \mathcal{L^+}$:
 \begin{align}
\tag{Part-1} \label{partpr1} \pr{\top\vert \Gamma}=1 & & \pr{\bot\vert \Gamma}=0\\
\tag{Part-2} \label{partpr2} \varphi \in \Gamma \Rightarrow \df{\varphi\vert \Gamma} \\
  \tag{Part-3} \label{partpr3} \df{\varphi\vert \Gamma} \Leftrightarrow \df{\n \varphi\vert \Gamma}  & &  \df{\varphi \et \psi\vert \Gamma} \Leftrightarrow \df{\psi \et \varphi\vert \Gamma}  \\
  \tag{Part-4}\label{partpr4}\pr{\varphi \et \psi\vert \Gamma}>0 \Rightarrow \df{\varphi\vert \Gamma}, \df{\psi, \vert \Gamma} & & \pr{\varphi\vert \Gamma}=0 \Rightarrow \df{\varphi \et \psi\vert \Gamma}\\
 \tag{Part-5}\label{partpr5} \ndf{\varphi\vert \Gamma} \Rightarrow \ndf{\varphi \et \psi\vert \Gamma} & & \mbox{ unless } \pr{\psi\vert \Gamma}=0 \\
 \tag{Part-6}\label{partpr6} \mbox{If  } \pr{\varphi\vert \Gamma}>0, \pr{\varphi \et \psi\vert \Gamma}=0, & & \mbox{then }\pr{\psi\vert \Gamma}=0 
 \end{align}

 \eqref{partpr1} requires that logical truths  have probability 1 and logical contradictions always have probability 0. \eqref{partpr2} requires that the probability of a claim given a set of premises that includes it is always defined. \eqref{partpr3} states that the conditional probability of a claim is defined just in case the conditional probability of its negation is, and that the order of conjuncts has no impact on whether the conditional probability of a conjunction is defined. According to \eqref{partpr4}, the conditional non-zero probability of a conjunction is defined  only if the conditional probability of both conjuncts is. Moreover, if the conditional probability of a conjunct is 0, the conditional probability of the conjunction is defined (and by the fact that the credence has an extension to a total conditional probability satisfying the standard axioms, we also know that it will be 0 as well). \eqref{partpr5} says that, unless this unusual circumstance occurs, the conditional probability of a conjunction is undefined if the conditional probability of at least one conjunct is. Finally, \eqref{partpr6} demands that if a conjunct has a conditional probability $>0$, then the conjunction has conditional probability 0 only if the other conjunct does.

Since the fact-finders are supposed not to be biased and aren't informed about the trial yet, we additionally assume that the priors of guilt, of what the evidence and what the narrations are, are undefined: $\ndf{G}$, $\ndf{g_1\et \cdots \et g_l}$, $\ndf{E\varphi}, \ndf{N_i\varphi}$ for any $\varphi\in \mathcal{L}^+$, and any $1<i<k$.

Four types of stances that a fact-finder might take towards a claim will be considered. First, a fact-finder might consider a claim completely uncontroversial, and accept it without any further argument.  Such stance will be modeled by the credence in a given claim reaching the \emph{uncontroversial acceptability threshold}, $\mathtt{a}$. On the opposite side of spectrum we have the \emph{negligibility threshold}, $\mathtt{n}=_{df} 1-\mathtt{a}$.  Notice that $\mathtt{a}$ and $\mathtt{n}$ can't be respectively 1 and 0, for this would require complete unrevisable certainty. 

One more type of stance needs to be incorporated into the framework --- that of \emph{strong plausibility}. Usually there are claims that are strongly supported while not being as close to certainty as the uncontroversially acceptable ones.  This  kind of credence that we would normally find sufficient for acting upon in our uncertain world will be denoted by $\mathtt{s}$. The opposite will be called \emph{rejectability}, $\mathtt{r}=_{df} 1-\mathtt{s}$.   Clearly we should require $\mathtt{a}>\mathtt{s}> \mathtt{r}>\mathtt{n}$.

\subsection{Information and updates}

After all the evidence and all the arguments have been presented in court, the background knowledge obtained now consists of  the  pieces of evidence presented, $\mathtt{E}$,  information about what is not part of evidence, $\mathtt{E}^-$,   
 the content of the guilt statement $\mathtt{G}$, and the description of the content of a certain finite assembly of finite theories meant to defend or accuse the defendant, $\mathtt{N}^D_1,\dots, \mathtt{N}^D_k, \mathtt{N}^A_1, \mathtt{N} ^A_m$ (and what is not part of which narration).  When making various assessments in the fact-finding process, at various stages one needs to conditionalize on various parts of the available information, depending on what is being assessed. The variants used are listed in the table below.
 \begin{center}
 \begin{tabular}{|p{3.2cm}|l|l|}
 \hline \hline
 \textbf{name} & \textbf{notation} & \textbf{meaning}\\ \hline 
 \footnotesize full &
  \footnotesize  $\mathtt{P}^f(\varphi \vert \Gamma)$ &
   \footnotesize $\mathtt{P}(\varphi\vert  \mathtt{E}, \mathtt{E}^d, \mathtt{E}^-,\mathtt{N}^d, \mathtt{N}^-, \mathtt{G}, \Gamma)$\\ 
   \footnotesize n-full &
     \footnotesize  $\mathtt{P}^{nf}(\varphi \vert \Gamma)$ &
      \footnotesize $\mathtt{P}(\varphi\vert  \mathtt{E}, \mathtt{E}^d,\mathtt{N}^d, \mathtt{N}^-, \mathtt{G}, \Gamma)$\\ 
 \footnotesize informed &
  \footnotesize $\mathtt{P}^i(\varphi \vert \Gamma)$ &  \footnotesize $\mathtt{P}(\varphi\vert  \mathtt{E}, \mathtt{E}^d, \mathtt{N}^d, \mathtt{G}, \Gamma)$\\ 
  \footnotesize evidential & 
   \footnotesize $\mathtt{P}^e(\varphi \vert \Gamma)$ &  \footnotesize $\mathtt{P}(\varphi\vert   \mathtt{E}, \mathtt{E}^d, \mathtt{E}^-, \mathtt{G}, \Gamma)$\\ 
 \footnotesize argued & 
  \footnotesize $\mathtt{P}^a(\varphi \vert \Gamma)$ &  \footnotesize $\mathtt{P}(\varphi\vert   \mathtt{N}^d, \mathtt{G}, \Gamma)$\\ 
   \footnotesize play-along &
    \footnotesize  $\mathtt{P}^{N_j}(\varphi \vert \Gamma)$ &  \footnotesize $\mathtt{P}(\varphi\vert  \mathtt{N}_j, \mathtt{N}^d, \mathtt{N}^-, \mathtt{G}, \Gamma)$\\ 
   \footnotesize n-extended play-along &  \footnotesize $\mathtt{P}^{nN_j}(\varphi \vert \Gamma)$ &  \footnotesize $\mathtt{P}(\varphi\vert  \mathtt{N}_j, \mathtt{E}, \mathtt{E}^d, \mathtt{N}^d, \mathtt{N}^-, \mathtt{G}, \Gamma)$\\ 
  \footnotesize e-extended play-along &  \footnotesize $\mathtt{P}^{eN_j}(\varphi \vert \Gamma)$ &  \footnotesize $\mathtt{P}(\varphi\vert  \mathtt{N}_j, \mathtt{E}, \mathtt{E}^d, \mathtt{E}^-, \mathtt{N}^d,  \mathtt{G}, \Gamma)$\\ 
    \footnotesize f-extended play-along &  \footnotesize $\mathtt{P}^{fN_j}(\varphi \vert \Gamma)$ &  \footnotesize $\mathtt{P}(\varphi\vert  \mathtt{N}_j, \mathtt{E}, \mathtt{E}^d, \mathtt{E}^-, \mathtt{N}^d,  \mathtt{N}^-, \mathtt{G}, \Gamma)$\\ 
\hline \hline
 \end{tabular}
  \end{center}

\section{Defining conditions on a set of  narrations}

Think of $\mathtt{N}$ as the set of attacking and defending narrations  seriously considered in the fact-finding process. In this section I list some basic conditions on a set of sets of sentences to count as a set of narration. In the next section, I explicate the requirements used in the evaluation of narrations.
\begin{align}
\tag{Exclusion} \label{Exclusion} \mathtt{P}^f(\n( \mathtt{N}_i \wedge \mathtt{N}_j))\geq \mathtt{a},  \mbox{ for }  i\neq j\\
\tag{Decision} \label{Decision} \mathtt{P}^{f\mathtt{N}^A_i}(G)\geq \mathtt{a} \et \mathtt{P}^{f\mathtt{N}^D_k}(\n G)\geq \mathtt{a} \\
\tag{Initial plausibility} \label{Initial plausibility} \mathtt{P}^e(\mathtt{N}_k)\geq \mathtt{n}  \\
\tag{Exhaustion} \label{Exhaustion} \mathtt{P}^f( \mathtt{N}_1\vee \cdots \vee  \mathtt{N_k}) \geq s 
\end{align}

\eqref{Exclusion}  requires  that  narrations under consideration should pairwise exclude each other  given what we know about the case.
According to \eqref{Decision}, a defense narration should clearly state that, given all that is known, the accused is not guilty, and an accusing narration should clearly state that, given all that is known, they are. \eqref{Initial plausibility} says that  we shouldn't consider narrations that are uncontroversially excluded by sensible background knowledge or by evidence. \eqref{Exhaustion} requires   that it should be strongly plausible that at least one of the narrations hold, given all that we know about the case.

\section{Evaluation criteria}

\paragraph{Explaining evidence.} Now we're ready to look at the explication of the criteria involved in the evaluation of competing narratives. Let's start with the requirement that they should explain the evidence. After all the narrations have been presented and deployed, an accusing narration $\mathtt{N}^A_i$ should ``make sense'' of evidence in the following sense. For any item of evidence presented, $e$, if,  according to
 $\mathtt{N}^A_i$,  it is not excluded  as evidence,   it  should be strongly plausible given $\mathtt{N}^A_i$.
\begin{align}
\tag{Explaining evidence A}
\label{Explaining evidence A}
 \mbox{For any } e\in \mathtt{E}, [  \n  \mathtt{P}^{\mathtt{N}^A_i}(\n E e) \geq \mathtt{s} \Rightarrow \mathtt{P}^{\mathtt{N}^A_i}(e) \geq \mathtt{s}  ] 
\end{align}

The sense in which a defending narration is supposed to explain evidence is somewhat different. After all, if the defense story is rather minimal and mostly constitutes in rebutting the accusations, it isn't reasonable to expect the defense to explain all pieces of evidence, as long as they aren't really used to support the opposing accusing narration.  Rather, the defense should argue that the possibility of the evidence being as it is while the defense's narration is true hasn't been rejected. Thus, we put the condition on a defending narration $\mathtt{N}^D_k$ as follows:
\begin{align}
\tag{Explaining evidence D}\label{Explaining evidence D} 
\mbox{For any } e\in \mathtt{E}, \mbox{ if there is } N^A_i \\
\nonumber
\mbox{such that } \mathtt{P}(\mathtt{N}^A_i\vert e)> \mathtt{P}(\mathtt{N}^A_i),
\\
\nonumber
\mbox{then } \mathtt{P}^{\mathtt{N}^D_k}(e)\geq \mathtt{r}.
\end{align}

\paragraph{Missing evidence.}

 The intuition here is that sometimes, given the  narration and whatever evidence we already have, certain evidence should be available, but it isn't. For instance, in a drunk driving case  the fact-finders would naturally expect a breathalyzer result, and in a murder case evidence as to how the victim was killed is needed.
    \begin{align}
  \tag{Missing evidence}
  \label{Missing evidence}
\mathbf{ME}(\mathtt{N}_i)  \Leftrightarrow & \mbox{ for some } \varphi_1, \dots, \varphi_u  \not \in \mathtt{E}: \\ \nonumber
&  [\mathtt{P}^{nN_i}(E(\varphi_1)\vee \cdots \vee E( \varphi_u)) \geq \mathtt{s}  ]  \end{align}

The disjunction above is there to ensure generality: it might be the case that some evidence from among a group of possible pieces of evidence would be needed, without any particular piece of evidence being expected. 

\paragraph{Gaps.}
 Sometimes a narration should be more specific, given what it says and what we already know. For instance, an accusing narration might be required to specify how the victim was attacked, or a defending narration should specify where the defendant was at the time of the crime. 
 Accordingly, we say that $\mathtt{N}_i$ is \textbf{gappy}  ($\mathbf{G}(\mathtt{N}_i)$)  just in case there are claims that the narration should choose from and yet it doesn't:
\begin{align}
\tag{Gap} \label{Gap}
\mathbf{G}(\mathtt{N}_i)  \Leftrightarrow & \mbox{ for some } \varphi_1, \dots, \varphi_u \not \in \mathtt{N}_i\\ \nonumber
 &  \mathtt{P}^{f\mathtt{N}_i}(\varphi_1 \vee\cdots \vee \varphi_u)\geq \mathtt{s} \wedge\\ \nonumber
& \mathtt{P}^{eN_i}(N_i(\varphi_1)\vee \cdots \vee N_i(\varphi_u))\geq \mathtt{s} 
\end{align}

\paragraph{Dominating accusing narration.} An accusing narration $\mathtt{N}^A_i$ \emph{dominates} the set of all accusing narrations $\mathbb{N}^A$ just in case it doesn't miss any evidence, it doesn't contain any gap,  in light of all available information and evidence  it is at least as likely any  other accusing narration, and it is strongly plausible, given all available information:
\begin{align}
\tag{Domination}
\label{Domination}
\mathbf{D}(\mathtt{N}^A_i) \Leftrightarrow  & \n \mathbf{ME}(\mathtt{N}^A_i) \et \n \mathbf{G}(\mathtt{N}^A_i) \et \\ \nonumber
& \mathtt{P}^f( \mathtt{N}^A_i)\geq \mathtt{P}^f( \mathtt{N}^A_j ) \mbox{ for all } j \neq i \et\\  \nonumber 
& \mathtt{P}^f(\mathtt{N}^A_i) \geq \mathtt{s}
\end{align}

\paragraph{Resiliency.}   A dominating narration $\mathtt{N}^A_i$ is \emph{resilient} ($\mathbf{R}(\mathtt{N}^A_i)$) just in case there is no non-negligible potential evidence that might undermine it, at least  in light of all we know (minus the negative description of the evidence, to avoid triviality)  ---  that is, no $\varphi$ with $\mathtt{P}^{nf}(E\varphi)\geq \mathtt{n}$ -- such that if $\mathtt{E}$ was modified to $\mathtt{E}\cup\{\varphi\}$, $\mathtt{N}^A_i$ would no longer dominate.

  \paragraph{Conviction beyond reasonable doubt.}
  A defense narration $\mathtt{N}^D_k$ \emph{raises reasonable doubt} ($\mathbf{RD}(\mathtt{N}^D_k)$) if it  has no gaps, and hasn't been rejected given all that we know:
  \begin{align}
  \tag{Reasonable doubt} \label{Reasonable doubt} 
  \mathbf{RD}(\mathtt{N}^D_k) \Leftrightarrow & \n \mathbf{G}(\mathtt{N}^D_k)   \et \mathtt{P}^f(\mathtt{N}^D_k) \geq \mathtt{r}
  \end{align}
  Accordingly,  we say that a conviction is \emph{beyond reasonable doubt} if it is justified by a resilient dominating narration and no defense narration raises reasonable doubt.

\section{Looking at the gatecrasher paradox}

Once we've formulated the framework, it will be instructive to use it to look at the gatecrasher paradox, and to observe how the requirements and distinctions introduced help us obtain better insight. 

Let's creatively call the suspects 1, 2, 3, \dots, 1000. Consider the situation in which the accusing narration is simply \emph{$1$ gatecrashed}, $\mathtt{g}_1$, and the defending narration is simply \emph{1 didn't gatecrash}, $\n \mathtt{g}_1$. Suppose we  have $\prcon{g_1}{e}{}=999/1000$ and $\prcon{\n g_1}{e}{}=1/1000$.

\begin{center}\begin{tabular}{|ll|}
\hline 
\footnotesize \textbf{Variant 1} &
\footnotesize $\mathtt{N}^A=\mathtt{g}_1$,  $\mathtt{N}^D=\n \mathtt{g}_1$. 
\\ \hline
\end{tabular}\end{center}
One might try to counter this formulation with the following strategy:
\begin{center}\begin{tabular}{|lp{6cm}|}
\hline \footnotesize \textbf{Strategy 1: Extreme thresholds}&
\footnotesize Take $\mathtt{P}^f(\cdot)$ to be  $\mathtt{P}(\cdot \vert \mathtt{e})$, and  claim $\mathtt{r}<1/1000$ and $\mathtt{s}>999/1000$.\\
\hline 
\end{tabular}\end{center}
The strategy isn't too successful, though --- it's \emph{ad hoc} and it isn't  immune to other versions of the paradox, where the number of people is tweaked to get the guilt probability threshold about whatever threshold was picked in Strategy 1. Perhaps, a better approach would be this:
\begin{center}
\begin{tabular}{|lp{6cm}|}\hline \footnotesize \textbf{Strategy 2: Full credence vs. statistics}&
\footnotesize Take $\mathtt{P}^f(\cdot)$ to \textbf{not} be  $\mathtt{P}(\cdot \vert \mathtt{e})$, and  claim  $\mathtt{P}^f(g_1)$ is insufficiently high. \\ \hline
\end{tabular}
\end{center}

Perhaps there is something to saying that the posterior credences don't have to match the statistical probabilities (as Kaye \cite{kaye1979paradox} suggests). But merely saying so doesn't provide a principled explanation: the fact that we're not inclined to accept a claim that has high statistical probability is the \emph{explanandum}, not the \emph{explanans}.

\begin{center}
\begin{tabular}{|lp{7cm}|}\hline \footnotesize \textbf{Strategy 3:  realistic gaps and non-resiliency}  &
There are multiple natural questions that the accusing narration fails to answer; it also fails to ensure that no future evidence is likely which might overturn the decision.\\ \hline
\end{tabular}
\end{center}

This is, pretty much, the strategy pursued in \cite{di2013statistics}. While it is a perfectly valid strategy if the question is how a case like gatecrasher would be handled in reality, this is not the end of the story.

The gatecrasher, or at least a variant thereof, can be viewed rather as an abstract thought experiment formulated to make a conceptual point. And when a philosopher formulates an abstract thought experiment, they're free to set it up any way they want, without too much attention paid to the level of realism, as long as appropriate conceptual restraints are satisfied.

 In particular, one might simply make it part of the description of the situation that no further evidence can be obtained despite everyone's best attempts, and no realistic considerations about further details can play a role. No witnesses can come and testify as to the character of 1, because he's been living under a rock and knows nobody. No question about the exact location  is natural, because the  stadium is somewhat unusual in having no gates, and in fact the 999 people who didn't pay for tickets jumped the fence, in an evenly distributed manner. No surveillance was possible because an accident in the local atomic plant fried all electronics in the area, etc. Call such a variant of the paradox \emph{abstract gatecrasher}.

The question now is: putting the issues with realism and supposing the abstract gatecrasher is immune to these, can we think of more purely epistemic or deontic reasons why the accusing narration in the abstract gatecrasher is not sufficient for a conviction beyond reasonable doubt?

  One issue that may come to mind even from a more abstract perspective, is that it doesn't seem that the accusing narration $\mathtt{N}_1=\mathtt{g}_1$ explains the evidence.

\begin{center}
\begin{tabular}{|lp{7cm}|}\hline \footnotesize \textbf{Strategy 5: Unexplained evidence} &\footnotesize 
The narration relies on $\mathtt{e}$ without entailing  that it shouldn't be evidence, and yet $\mathtt{P}^{N_1}(\mathtt{e} ) \not \geq \mathtt{s}$. \\ \hline
\end{tabular}
\end{center}

 The mere fact that 1 decided to gatecrash, while suggesting that he might've done it with a bunch of friends, certainly doesn't make the claim that so did other 998 people strongly plausible.

This, however, can be fairly easily fixed by modifying the accusing narration to \emph{1 gatecrashed, and so did 998 other people}. This, formally comes to using 
\begin{center}\begin{tabular}{|ll|}
\hline 
\footnotesize \textbf{Variant 2} &
$\mathtt{N}'_1=\{\mathtt{g}_1, \mathtt{e}\}$ \\
\hline\end{tabular}\end{center}
 Indeed, the mere fact that 1 decided to gatecrash, while suggesting that he might've done it with a bunch of friends, certainly doesn't make the claim that so did other 998 people strongly plausible. Quite clearly, 
$\mathtt{P}^{N_1}(e\vert  \mathtt{N}'_1 ) \geq \mathtt{s}$, and so Strategy 5 doesn't beat the paradox with this narration as the accusing narration. Can we do any better?

\section{Epistemological commitment issues}

Imagine the sides have the following exchange:
\begin{center}\begin{tabular}{lp{10cm}}
Defense: & So you claim that the suspect is responsible for the damage?\\
Persecution: & Yes.\\
D: & And you agree that the claim that the sidewalk was unusually slippery that day due to an oil truck failure is at least as likely as that the suspect is responsible? \\
P: & Of course. \\
D: & And, given that you're accusing my client, you think you're in position to claim you know he is responsible, correct?\\
P: & Yes, that's correct.\\
D: & Are you not, then, in position to know that  the sidewalk was unusually slippery that day due to an oil truck failure?\\
P: & $\dots$
\end{tabular}
\end{center}

I hope the reader shares the intuition that responding ``no'' to the above question would be a sign of a serious cognitive failure.  In general, it seems that we'd normally expect the persecution to accept any claim relevant to the case that, given the evidence and the persecution's narration, is at least as likely as the guilt statement that they're putting forward. This motivates the following requirement:

\begin{center}
\begin{tabular}{lp{9cm}}
(Commitment) & For any $\varphi$ relevant to the case, if   $\prtext{\varphi}{f\mathtt{N}^A_i}\geq \prtext{G}{f\mathtt{N}^A_i} $, then $\mathtt{P}^{eN_i}(N_i(\varphi))\geq \mathtt{s}$.
\end{tabular}
\end{center}

If the reader prefers to do so, we might leave the notion of relevance at the intuitivel level. We don't have to though --- it could be explicated along the following lines. First, a set of sentences is relevant for the case if  it is consistent with the background knowledge and there is a narration such that its posterior probability given all background knowledge together with that set is different from its posterior probability given all background knowledge only. A set of sentences is a minimal relevant set if no proper subset thereof is a relevant set. A sentence is relevant if it or its negation is a member of a minimal relevant subset.

Now, how does (Commitment) apply to the gatecrasher? Take any $\mathtt{g}_i$ where $i\neq 1$.  Consider an argument analogous to the previous one:

\begin{center}\begin{tabular}{lp{9cm}}
Defense: & So you claim that 1 is guilty?\\
Persecution: & Yes.\\
D: & And you agree 2 is at least as likely to be guilty as 1? \\
P: & Of course. \\
D: & So you're in position to claim you know 1 is guilty, correct?\\
P: & Yes, that's correct.\\
D: & Are you not, then, in position to know that 2 is guilty as well?\\
P: & $\dots$
\end{tabular}
\end{center}

Again, it seems that the negative answer to the above question would be irrational. (If you worry about $\mathtt{g}_i$ being relevant, wait for the development of the argument.)

Observe now that in the case of the gatecrasher, the accusing narration already discussed, $\mathtt{N}'_1$ is in the somewhat unusual lottery-paradox-like situation that given all that is known (and the narration itself) any other suspect is at least as likely to be guilty as suspect 1. 

Now we can run the argument to the effect that $\mathtt{N}'_1$ fails to satisfy conditions for conviction beyond reasonable doubt.
Given the relevance of all $g_i$ ($i\neq 1$), (Commitment) entails that for any $i\neq 1$ we have
\begin{align}
 \tag{Step 1} \label{Step 1}
 \mathtt{P}^{e{\mathtt{N}'}_1}({\mathtt{N}'}_1(\mathtt{g}_i))\geq \mathtt{s}. 
\end{align} 
Since $\mathtt{N}'_1$ is an accusing narration delivering 
 $\mathtt{g}_1$,  by \eqref{Decision}  and the fact that $\mathtt{a}>\mathtt{s}$ we  have  
\begin{align}
\tag{Step 2} \label{Step 2}
\mathtt{P}^{f\mathtt{N'}_1}(\mathtt{g}_i)\geq \mathtt{a}> \mathtt{s}.
\end{align}
The very description of $\mathtt{N}'_1$ entails:
\begin{align}
\tag{Step 3} \label{Step 3}
\mathtt{g}_i \not \in \mathtt{N}'_1.
\end{align}
Steps (1-3) taken together, however, constitute the defining elements of  \eqref{Gap} (for the slightly degenerate case where $\varphi_1 \vee \cdots \vee \varphi_u$ simply is $\mathtt{g}_i$). This means that $\mathtt{N}'_1$ is gappy, $\mathbf{G}(\mathtt{N}'_1)$, and so it fails to justify a conviction beyond reasonable doubt.

What happens, on the other hand, if $\mathtt{N}'_1$ is replaced with $\mathtt{N}^+_1$: an accusing narration which results with taking the least narration containing $\mathtt{N}'_1$ and closed under (Commitment) with respect to all $\mathtt{g}_i$, $i\neq 1$, so that:
\begin{align}
\tag{Bite the bullet}
\label{Bite the bullet} \mathtt{N}^+_1 = \{\mathtt{g}_1, \mathtt{e}\}\cup \{ \mathtt{g}_k\vert k\neq 1\}?
\end{align}
Then, the resulting narration simply becomes highly implausible: $\mathtt{e}$, and therefore also $\mathtt{N}^+_1$, entails that exactly one person is innocent, but also for each particular suspect $u$, $\mathtt{N}^+_1$ insists on $u$ being guilty. Such a narration doesn't  even  satisfy \eqref{Initial plausibility}, not to mention failing to be a dominating one (for which strong plausibility is required).

By the way, this is why each $\mathtt{g}_i$ is relevant: together they change the outcome. This is also the reason why they're relevant in the more technical sense: the set of all $\mathtt{g}_i$s contradicts evidence, while no proper subset thereof does. 

So, to wrap up the discussion of the gatecrasher, in a sense there is no single simple diagnosis of the gatecrasher, for the following reasons. First of all, there is no single gatecrasher, but two main variants thereof: a realistic one and a very abstract one.\footnote{To be honest, multiple variants thereof depending on how unrealistic we're required to be. But let's stay content with two extreme cases.} Second, some supposed solutions don't work against any of them. Third, sometimes there is  more than one reason why a variant fails. Four, different variants can be seen as failing for different reasons (although reasons that apply to the abstract one apply to the realistic one; it's just that deploying them to the realistic gatecrasher is an overkill).

\bibliographystyle{eptcs}

\end{document}